\documentclass{sig-alternate}

\usepackage{graphicx}
\usepackage{hyperref}
\usepackage[update,prepend]{epstopdf}
\usepackage{balance}
\usepackage{amsmath}
\usepackage{caption}
\usepackage{subcaption}
\usepackage{footnote}
\usepackage{bbding}

\usepackage{booktabs} 
\usepackage{multirow}
\usepackage{algorithm}
\usepackage{algorithmic}

\usepackage{amsfonts}
\usepackage{yyy-math}
\usepackage{textcase}
\usepackage{wrapfig}
\usepackage{graphicx}
\usepackage{xcolor}
\usepackage{subcaption}
\usepackage{epstopdf}
\usepackage{url}
\usepackage{tabularx}
\usepackage{paralist}
\usepackage{amssymb}
\usepackage{pifont}

\newtheorem{theorem}{Theorem}

\newtheorem{proposition}{Proposition}
\newtheorem{corollary}{Corollary}

\usepackage{listings}
\usepackage{color}
\definecolor{dkgreen}{rgb}{0,0.6,0}
\definecolor{gray}{rgb}{0.5,0.5,0.5}
\definecolor{mauve}{rgb}{0.58,0,0.82}
\newcommand*{\affmark}[1][*]{\textsuperscript{#1}}

\makeatletter
 \let\@copyrightspace\relax
\makeatother

\begin{document}

\title{Exploring High-Order Structure for Robust Graph Structure Learning}

\numberofauthors{5}
\author{
{Guangqian Yang\affmark[1], Yibing Zhan\affmark[3], Jinlong Li\affmark[1], Baosheng Yu\affmark[2], Liu Liu\affmark[2], Fengxiang He\affmark[3]}
\vspace{1.6mm} \\
\fontsize{10}{10}\selectfont\itshape
\affmark[1]University of Science and Technology of China\\
\fontsize{10}{10}\selectfont\itshape
\affmark[2]The University of Sydney\\
\fontsize{10}{10}\selectfont\itshape
\affmark[3]JD Explore Academy, JD.com\\
}

\maketitle
\begin{abstract}
Recent studies show that Graph Neural Networks (GNNs) are vulnerable to adversarial attack, i.e., an imperceptible structure perturbation can fool GNNs to make wrong predictions. Some researches explore specific properties of clean graphs such as the feature smoothness to defense the attack, but the analysis of it has not been well-studied. In this paper, we analyze the adversarial attack on graphs from the perspective of feature smoothness which further contributes to an efficient new adversarial defensive algorithm for GNNs. We discover that the effect of the high-order graph structure is a smoother filter for processing graph structures. Intuitively, the high-order graph structure denotes the path number between nodes, where larger number indicates closer connection, so it naturally contributes to defense the adversarial perturbation. Further, we propose a novel algorithm that incorporates the high-order structural information into the graph structure learning. We perform experiments on three popular benchmark datasets, Cora, Citeseer and Polblogs. Extensive experiments demonstrate the effectiveness of our method for defending against graph adversarial attacks.
\end{abstract}

\section{Introduction}
Graph Neural Networks (or GNNs) \cite{kipf2017semi, hamilton2017inductive, velickovic2018graph} play an important role in deep learning-based graph representation learning. By extending convolution operation to graph-structured data, graph neural networks show excellent performance in many applications, such as node classification~\cite{yang2016revisiting, kipf2017semi}, link prediction~\cite{grover2016node2vec, schlichtkrull2018modeling}, and graph classification~\cite{niepert2016learning, gilmer2017neural}.

Despite the success of GNNs in graph structure learning, recent studies have shown that GNNs are vulnerable to adversarial attacks, i.e., a small perturbation on the graph will lead to a drastic performance degradation on graph structure learning~\cite{szegedy2013intriguing, goodfellow2014explaining}. Specifically, by injecting imperceptible perturbation into node feature or graph structure~\cite{dai2018adversarial, zugner2018adversarial, zugner2018adversarial_, xu2019topology}, it can easily manipulate the prediction of GNNs. Therefore, the robustness of GNNs has received increasing attention from the community. Existing methods focusing on the robustness of GNNs can be divided into the following categories: adversarial training, robustness certification, and structure learning. This paper focus on the structure learning.

However, the low-order structure of the graph is vulnerable for defensing  adversarial attacks, and the structure learning-based methods aim to mitigate the impact of adversarial attacks and help GNNs learn the true distribution of graph structures \cite{zheng2020robust, jin2020graph, zhang2020gnnguard}. Compared to the initial structure, the high-order graph structure, which is reflected in the powers of the adjacency matrix, is naturally a more robust structure. Although adversarial attacks may perturb the low-order graph structures, they could hardly affect the high-order graph structures as the attacks can only change a small fraction of paths within the high-order graphs. As a result, the high-order graph structure information can naturally act as a denoising filter to make the low-order graph feature smoother.

Inspired by this, we propose a novel method that incorporates the high-order structural information into the learning process for robust graph structure learning. In this paper, we first analyze the graph adversarial attack from the perspective of graph feature smoothness, which is defined as the distance between connected nodes. And we both theoretically and empirically show that the adversarial structure perturbation essentially increases the local feature smoothness. We then devise a novel method to explore the high-order structural information for graph structure learning. Intuitively, the high-order adjacency matrix reflects the common neighbors between two nodes that could guide the structure learning. Though the adversarial attack may perturb some graph edges, it's less likely to have much perturbation on the overall distribution of high-order graphs, and thus the high-order structure can be used to alleviate the influence of adversarial perturbations. Furthermore, we also theoretically show that high-order graph is very effective in smoothing the graph and eliminating the influence of adversarial perturbations. Our main contributions are as follows:

\begin{itemize}
	\item We analyze the graph adversarial attack from the perspective of feature smoothness, i.e., high-order graph structure is a smoother filter whose overall distribution is less influenced by the adversarial attack.
	\item We explore the high-order graph structure to alleviate the influence of adversarial attack, which can be formulated as the normalized adjacency matrix regularization to guide the graph structure learning.
	\item We conduct extensive experiments on several popular datasets using different types of attacks, and analyze the performance and sensitivity under different attack settings. Experimental results demonstrate that our method is a universal method to defense graph adversarial attacks.
\end{itemize}

\section{Related Work}

In this section, we briefly review recent works on graph neural networks and adversarial attacks or defense for graph neural networks.

\subsection{Graph Neural Networks.} Graph Neural Networks are deep learning based methods for processing graph-structured data, and have shown excellent performance in many realistic tasks~\cite{yang2016revisiting, grover2016node2vec, niepert2016learning}. Generally Graph Neural Networks could be classified into spectral and spacial methods. The spectral methods generally learn graph filters based on graph spectral theory~\cite{estrach2014spectral} first generalize Convolutional Neural Networks to graph signals based hierarchical clustering and graph Laplacian. ChebNet~\cite{defferrard2016convolutional} utilizes Chebyshev polynomials as the fast localized spectral filter for computational efficiency. Graph Convolution Network \cite{kipf2017semi} exploits a localized first-order approximation of spectral filters to further simply the filtering operation. The spatial methods directly propagate information based message passing in spatial domain. GraphSAGE \cite{hamilton2017inductive} proposes an inductive learning framework on graphs that generates embeddings by sampling and aggregating neighbor information. Graph Attention Network~\cite{velickovic2018graph} proposes to apply attention mechanism on graph so as to learn different aggregation weight for neighbors based on their dependency. There are also many state-of-the-art methods recently.

\subsection{Adversarial Attack and Defense.} Deep learning models have been shown to be vulnerable to adversarial perturbation \cite{szegedy2013intriguing, goodfellow2014explaining}, and so as Graph Neural Networks. A large amount researches have focus on the adversarial attack and defenses recently\cite{zheng2021graph, sun2018adversarial, he2020robustness}. Generally, the attack models could be categorized into black box attack and white box attack based on the information the attacker could access about the model. Typical attack methods are as follows, where Nettack \cite{zugner2018adversarial} derives an incremental attack method which utilizes approximation, RL-S2V \cite{dai2018adversarial} uses Q-learning to add or drop edges from the graph sequentially, NIPA \cite{sun2020non} proposes a more practical node injection attack, GF-Attack \cite{chang2020restricted} aims to attack the graph filter of given models, Metattack \cite{zugner2018adversarial} treats the graph structure matrix as a hyperparameter to learn, IG-JSMA \cite{wu2019adversarial} introduces integrated gradients \cite{sundararajan2017axiomatic} based methods, PGD \cite{xu2019topology} utilizes projected gradient descent from a perspective of first-order optimization.

At the other end of the scale, defense methods aim to eliminate the influence of adversarial attack as much as possible. Based on the design principle, the defense models could generally be divided into certificate methods, adversarial training, and structure learning. The general idea of certificate methods is to ensure the prediction of the model not change in a certified perturbation radius, typical methods include attribute oriented \cite{zugner2019certifiable}, structure oriented \cite{bojchevski2019certifiable}, sparsity-aware certificate \cite{bojchevski2020efficient}, collective certificate \cite{schuchardt2020collective}, and so on. Adversarial training aims to directly improves the robustness of the model by training with adversarial examples, typical methods include RAWEN \cite{ding2020improving}, DWNS \cite{dai2019adversarial}. And Structure learning methods aim to learn graph structure from the perturbed graph, typical methods include NeuralSparse \cite{zheng2020robust} that learns to remove potentially task-irrelevant edges from input graphs, Pro-GNN \cite{jin2020graph} that explores graph properties of sparsity, low rank and feature smoothness, and GNNGUARD \cite{zhang2020gnnguard} that exploits the relationship between the graph structure and node features to mitigate negative effects of attack.

\begin{figure*}[!ht]
	\centering
	\includegraphics[width=1.0\linewidth]{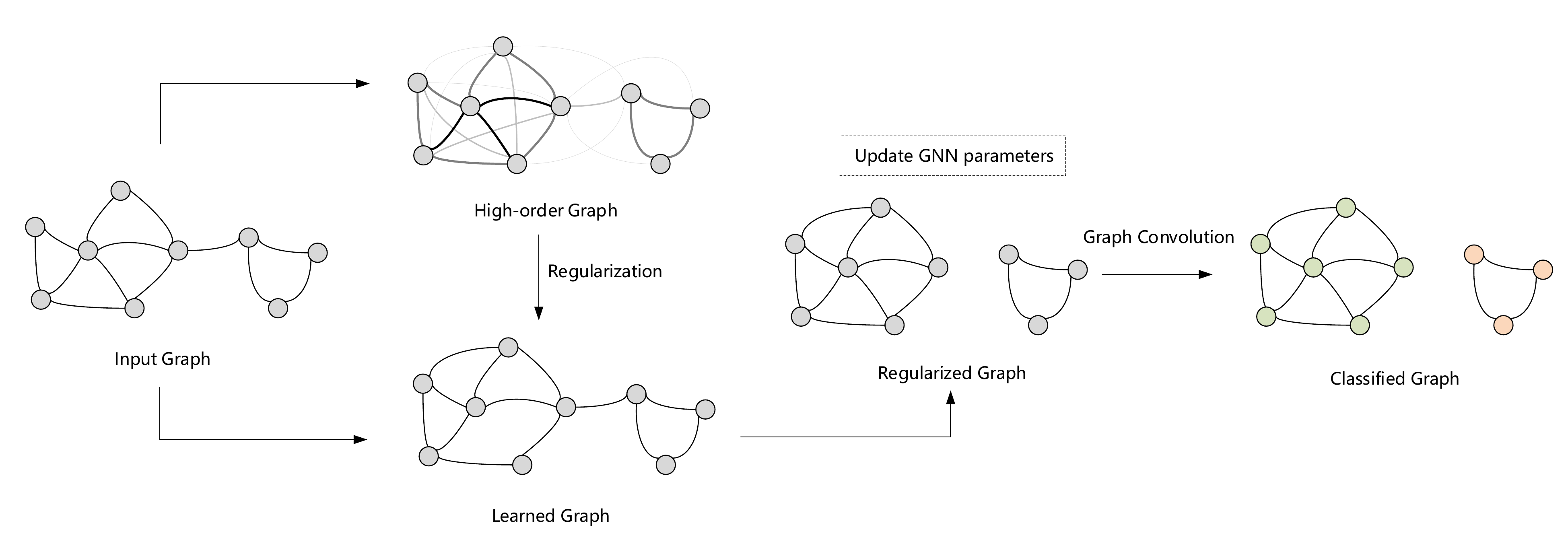}
	\caption{The work-flow of the our method. We use the initial input graph (perturbed) to learn a estimated graph and to obtain the high-order graph. Here the high-order graph can be viewed as a weighted graph, where nodes with larger connectivity have closer links. The high-order graph can be further used to guide the graph structure learning process.}
	\label{framework}
\end{figure*}

\section{Methods}

In this section, we first introduce the notations used in this paper. We then analyze how high-order structure helps defending adversarial attacks on graph. Lastly, we introduce the proposed high-order structure learning.

\subsection{Preliminaries}
We introduce the notations used in this paper as follows. Specifically, we use bold upper case letters for matrix, bold lower case letters for vector, and regular letters for scalar. Given a graph $\mathcal{G} = (\mathcal{V}, \mathcal{E})$, where $\mathcal{V}$ is the set of nodes, $\mathcal{E}$ is the set of edges, let $N = |\mathcal{V}|$ denote the number of nodes. For each node $v_i \in \mathcal{V}$, we have a corresponding attribute feature vector $\mathbf{x_i} \in \mathbb{R}^F$, and all nodes them form an attribute feature matrix $\mathbf{X} \in \mathbb{R}^{N\times F}$, where $F$ denotes the dimension of the attribute. In addition, the adjacency matrix of $\mathcal{G}$ is defined as $\mathbf{A} \in \mathbb{R}^{N\times N}$, where $\mathbf{A}_{ij} = 1$ indicates an edge between node $v_i$ and $v_j$, and $\mathbf{A}_{ij} = 0$ indicates no edge. For the node classification task, GNNs can be seen a function with parameters $\mathbf\theta$ that map the nodes into different classes, using the adjacency matrix $\mathbf{A}$ and the feature matrix $\mathbf{X}$ as inputs.
A general two-layer graph convolution network can be formulated as $f_\theta(\mathbf{A}, \mathbf{X}) = softmax(\mathbf{\hat A} \sigma(\mathbf{\hat AXW_1})\mathbf{W_2})$, where $\mathbf{\hat A}=\mathbf{\tilde{D}}^{-1/2}\mathbf{\tilde{A}}\mathbf{\tilde{D}}^{-1/2}$, $\mathbf{\tilde{A}}=\mathbf{A}+\mathbf{I}$ and $\mathbf{\tilde{D}}$ is a diagonal matrix where $\mathbf{\tilde{D}}_{ii}=\sum_j{\mathbf{\tilde{A}}_{ij}}$, 
$\mathbf{W_1}$ and $\mathbf{W_2}$ indicate trainable weight matrices, and $\sigma$ is the activation function such as ReLU~\cite{maas2013rectifier}.

Given an adjacency matrix $\mathbf{A}$ poisoned by the adversarial attack, our goal is to learn the estimated adjacency matrix $\mathbf{S} \in \mathbb{R}^{N\times N}$ that could alleviate the influence of attack, and the GNN parameters $\mathbf\theta$ for downstream tasks. Consistent with previous symbols, we use $\mathbf{\hat S}=\mathbf{\tilde{D}}^{-1/2}\mathbf{\tilde{S}}\mathbf{\tilde{D}}^{-1/2}$ to denote the normalized estimated adjacency matrix. The Laplacian of the graph is defined as $\mathbf{L} = \mathbf{\tilde{D}} - \mathbf{\tilde{A}}$.

\subsection{Adversarial Attacks Increase Smoothness}

As illustrated in previous work~\cite{zhu2021improving}, the attack losses increase only when removing a homophilous edge, or adding a heterophilous edge to node $v$. In this subsection, we analyze the adversarial attack problem from the perspective of feature smoothness. We both theoretically and empirically show that the above-mentioned attacks actually increase the feature smoothness of graph.

Similar to~\cite{jin2020graph}, we define the average graph feature smoothness as follows: 
\begin{equation}
	s = \frac{1}{\|\mathbf{\tilde{A}}\|_1} \sum_{i=1}^{N}\sum_{j=1}^{N}\mathbf{\tilde{A}}_{ij}(\mathbf{x}_i-\mathbf{x}_j)^2 =\frac{1}{\|\mathbf{\tilde{A}}\|_1}tr(\mathbf{X}^T\mathbf{L}\mathbf{X}),
	\label{feature smoothness}
\end{equation}
where the $\mathbf{x}_i$ and $\mathbf{x}_j$ are the feature of nodes $v_i$ and $v_j$, respectively. This formula utilizes the difference of node features to depict the overall graph smoothness, where larger distance indicates larger smoothness.

\begin{proposition}
Let $\mathcal{G} = (\mathcal{V}, \mathcal{E})$ be a graph with adjacency matrix $\mathbf{A}$ and feature matrix $\mathbf{X}$. The node features of $v_i$ is $\mathbf{x}_i = p\cdot onehot(y_i) + \frac{1-p}{|y|}\cdot \mathbf{1}$, where $\mathbf{1}$ is a vector with all elements equal to $1$, and $p$ is the uniform noise strength. We also assume a homophilous edge fraction $h$ of each neighbors that belongs to the same class and $h>\frac{1}{|\mathcal{Y}|}$ holds. The edge set $\mathcal{E}$ can be divided into two sets $\mathcal{E}_{homo}$ and $\mathcal{E}_{hetero}$, where $\mathcal{E}_{homo}\cup\mathcal{E}_{hetero}=\mathcal{E}$ and $\mathcal{E}_{homo}\cap\mathcal{E}_{hetero}=\emptyset$. The adversarial attack leads to the increase of feature smoothness.

\begin{proof}
	We consider a targeted attack, and the local smoothness is defined as:
	\begin{align}
			s_{i} &= \frac{1}{\|\mathbf{\tilde A}_i\|_1} \sum_{j=1}^{N}\mathbf{\tilde A}_{ij}(\mathbf{x}_i-\mathbf{x}_j)^2 \nonumber\\ 
			&= \frac{1}{\tilde d_i}(\tilde d_i\cdot h\cdot s_{1} + \tilde d_i\cdot (1-h)\cdot s_{2}),
		\label{local feature smoothness}
	\end{align}
	where $\tilde{d}_i = d_i+1$ is the degree of node $v_i$, $s_{1}=(\mathbf{x}_i-\mathbf{x}_j)^2=0$ where $e_{ij}\in\mathcal{E}_{homo}$, and $s_{2}=(\mathbf{x}_i-\mathbf{x}_j)^2=2p^2$ where $e_{ij}\in\mathcal{E}_{hetero}$.
	We then add adversarial perturbation on the graph. To simplify the graph without loss of generality, we only modify one edge to see how the smoothness changes as follows.
	
	\begin{itemize}
		\item\textbf{Add a heterophilous edge.} The local feature smoothness of node $v_i$ becomes:
		\begin{align*}
				s^\prime_{i} &= \frac{1}{\tilde d_i+1}(\tilde d_i\cdot h\cdot s_{1} + (\tilde d_i\cdot (1-h)+1)\cdot s_{2}) \\
				&= (1-\frac{\tilde d_i}{\tilde d_i+1}\cdot h)\cdot 2p^2 
				\\
				&> \frac{1}{\tilde d_i}(\tilde d_i\cdot h\cdot s_{1} + \tilde d_i\cdot (1-h)\cdot s_{2}) 
				= s_{i}.	
		\end{align*}
		\item\textbf{Delete a homophilous edge.}
		The local feature smoothness of node $v_i$ becomes:
		\begin{align*}
				s^\prime_{i} &= \frac{1}{\tilde d_i-1}((\tilde d_i\cdot h-1)\cdot s_{1} + \tilde d_i\cdot (1-h)\cdot s_{2}) \\
				&= \frac{\tilde d_i}{\tilde d_i-1} (1-h)\cdot 2p^2 \\
				&> \frac{1}{\tilde d_i}(\tilde d_i\cdot h\cdot s_{1} + \tilde d_i\cdot (1-h)\cdot s_{2}) 
				= s_{i}.
		\end{align*}
	\end{itemize}
	Therefore, we have that either adding a heterophilous edge or deleting a homophilous edge will increase of local feature smoothness.
\end{proof}
\end{proposition}

\begin{figure}[htbp]
	\centering
	\includegraphics[width=0.8\linewidth]{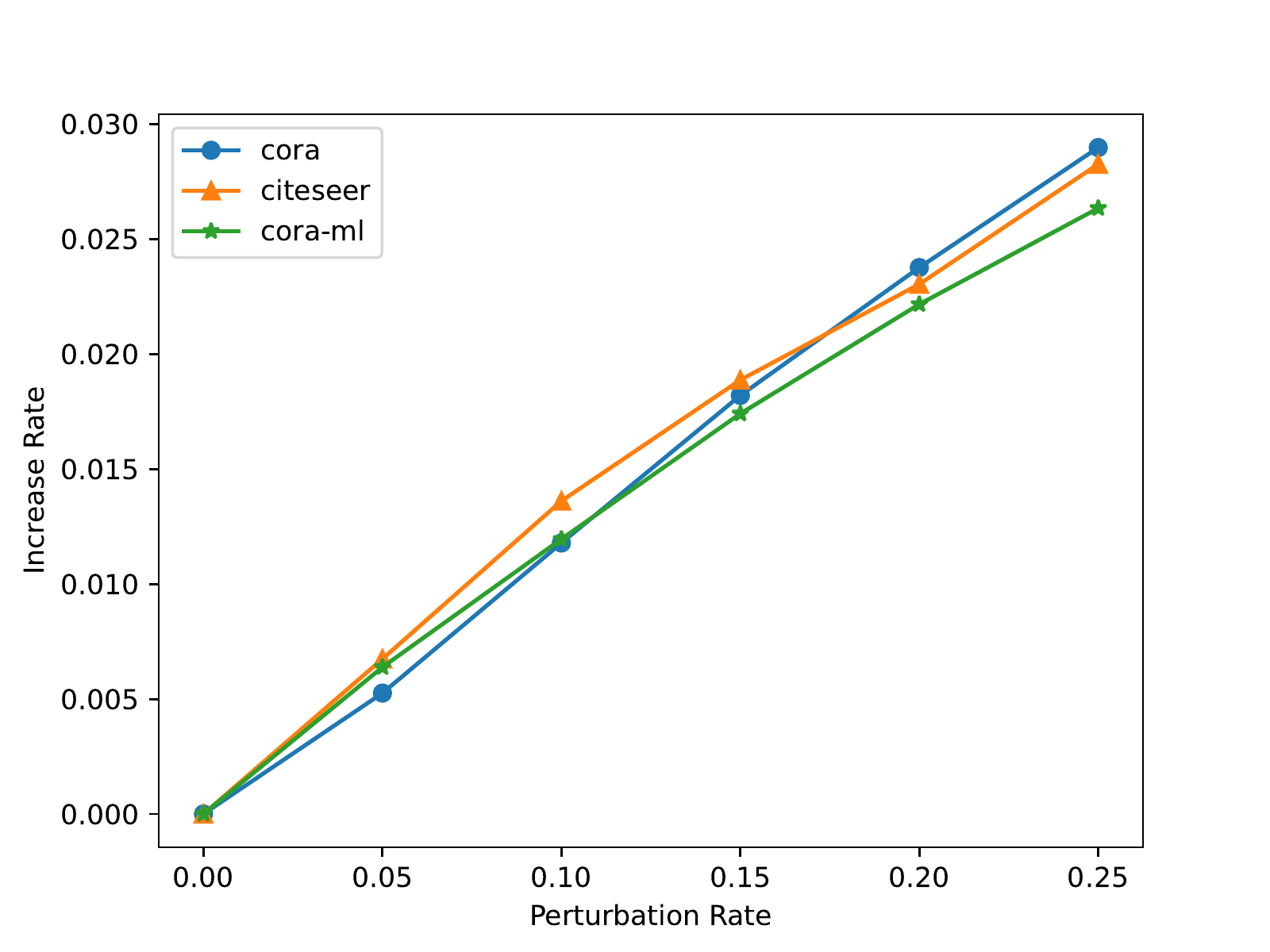}
	\caption{The influence of different perturbation rates on the smoothness.}
	\label{smoothness}
\end{figure}

\subsubsection{Empirical Evidence.} 
Intuitively, adding a heterophilous edge will cause the feature of node $v_i$ aggregated with features different from the node, which makes the feature move towards the features of different label to cause the misclassification, and deleting a homophilous edge is on the contrary. To validate our analysis, we conduct some experiments and calculate the graph smoothness by  $\frac{1}{\|\mathbf{A}\|_1}tr(\mathbf{X}^T\mathbf{L}\mathbf{X})$. Here we conduct attack experiments with different perturbation rate on several datasets.
As shown in Figure~\ref{smoothness}, we find that the feature smoothness increases when increasing the perturbation rate, which is consistent with our proposition. This phenomenon also indicates that the adversarial attack essentially increases the local smoothness of graph to fool the model into making wrong prediction.

\subsection{High-Order Graph is a Smooth Filter}

High-order graph indicates the powers of the normalized matrix. Intuitively, the perturbation of high-order structure is much less than the initial graph structure. We have the normalized graph Laplacian as follows:
\begin{equation}
	\mathbf{\hat{L}} = \mathbf{\tilde{D}}^{-1/2}(\mathbf{\tilde{D}}-\mathbf{\tilde{A}})\mathbf{\tilde{D}}^{-1/2} = \mathbf{I} - \mathbf{\hat{A}},
\end{equation}
where the symmetric graph Laplacian is positive semi-definite. 
\begin{theorem}
Let $\mathbf{\tilde{A}}=\mathbf{A}+\mathbf{I}$ be the adjacency matrix with self-loop. Denote the eigenvalues of $\mathbf{\tilde{A}}$ as $\lambda_1 \geq \lambda_2 \geq ... \geq \lambda_n$. For each eigenvalue $\lambda$, we have that the eigenvalues satisfy $|\lambda|\leq 1$, i.e., $|\lambda^k| \leq |\lambda|$ for any $k\geq1$.

\begin{proof} 
	Let $\mathbf{u}$ be any real vector of unit norm, then we have 
	\begin{align*}
			\mathbf{u}^T\mathbf{\hat{L}}\mathbf{u} &= \frac{1}{2}\sum_{i,j=1}^{N}\mathbf{\tilde{A}}_{ij}(\mathbf{\tilde{D}}^{-1/2}u_i-\mathbf{\tilde{D}}^{-1/2}u_j)^2 \\
			&\leq\sum_{i,j=1}^{N}\mathbf{\tilde{A}}_{ij}(\mathbf{\tilde{D}}^{-1}u_i^2+\mathbf{\tilde{D}}^{-1}u_j^2) = 2\mathbf{\tilde{A}}_{ij}u_i^2\mathbf{\tilde{D}}^{-1} \\
			&=2\sum_{i=1}^{N}u_i^2\mathbf{\tilde{D}}^{-1}\sum_{j=1}^{N}\mathbf{\tilde{A}}_{ij} = 2\sum_{i=1}^{N}u_i^2\mathbf{\tilde{D}}^{-1}\mathbf{\tilde{D}} \\
			& = 2\sum_{i=1}^{N}u_i^2 = 2.
	\end{align*}
	Therefore, we have that the largest eigenvalue of $\mathbf{\tilde{L}}$ is upper bounded by 2, i.e., the eigenvalue of $\mathbf{\tilde{A}}$ satisfy $|\lambda|\leq1$ and $|\lambda^k| \leq |\lambda|$ for any $k\geq1$. Note that the $0$ is the largest eigenvalue of $\mathbf{\tilde{L}}$, so $1$ is the eigenvalue of $\mathbf{\tilde{A}}$, and for the rest of the eigenvalues we have $|\lambda^k| \le |\lambda|$.
\end{proof}
\end{theorem}

From Theorem 1, we know that by taking powers $K>1$, the spectrum allows the filter to act as a low-pass type filter. This filter makes the frequencies of graph signal become lower, i.e., the local graph feature becomes smoother. 


\subsection{High-Order Structure Learning}

In this subsection, we explore the high-order normalized adjacency matrix for graph structure learning, which can naturally preserve the high-order structural similarity and smooth the graph. The high-order structure learning can be formulated as:
\begin{equation}
	\mathcal{L}_s = \eta_1\|\mathbf{\hat{S}} - \mathbf{\hat{A}} \|_F^2 + \sum_{k=2}^{K}\eta_k\|\mathbf{\hat{S}} - \mathbf{\hat{A}}^k \|_F^2,
	\label{eq4}
\end{equation}
where the first term $\eta_1\|\mathbf{\hat{S}} - \mathbf{\hat{A}} \|_F$ minimizes the difference between the learned adjacency matrix and the perturbed adjacency matrix. High-order terms are used to improve the structure learning. One notable difference between second- and third-order is the symbol of eigenvalue, where the second-order filter has only positive eigenvalues. Though the filter becomes smoother with $k$ becoming larger, it also leads to the over-smoothing problem~\cite{li2018deeper}, where the representations become too smooth to distinguish. In practice, we use $K=2$ or $K=3$.
From the geometric view, the element of high-order normalized adjacency matrix $\mathbf{\tilde{A}}^k_{ij}$ denotes the $k$-hop transition probability between two nodes $v_i$ and $v_j$, where node pairs with larger connectivity are assigned with larger weights. Intuitively, though the adversarial attack may add/delete a edge between node $v_i$ and $v_j$, it has less influence on the distribution of the high-order adjacency power. Therefore, exploring the high-order adjacency matrix can be helpful to alleviate the influence of adversarial attack.

\begin{corollary}
Let $\mathbf{L} = \mathbf{\tilde{D}} - \mathbf{\tilde{A}}$ be the graph Laplacian defined as $\mathbf{L} = \mathbf{\tilde{D}} - \mathbf{\tilde{A}}$, and $\mathbf{L}_{2} = \mathbf{\tilde{D}} - \mathbf{\tilde{A}}\mathbf{\tilde{D}}^{-1}\mathbf{\tilde{A}}$ be the 2-order graph Laplacian. For an identity feature matrix $\mathbf{X}$, we have $tr(\mathbf{X}^T(\mathbf{L}-\mathbf{L}_{2})\mathbf{X})>0$.

\begin{proof}
	According to the properties of Laplacian matrix, we have:
	\begin{align*}
			tr(\mathbf{L})-tr(\mathbf{L}_{2})&=tr(\mathbf{\tilde{D}}-\mathbf{\tilde{A}})-tr(\mathbf{\tilde{D}} - \mathbf{\tilde{A}}\mathbf{\tilde{D}}^{-1}\mathbf{\tilde{A}}) \\
			&=tr(\mathbf{\tilde{A}}\mathbf{\tilde{D}}^{-1}\mathbf{\tilde{A}}-\mathbf{\tilde{A}}) \\
			&=tr(\mathbf{\tilde{A}}\mathbf{\tilde{D}}^{-1}\mathbf{\tilde{A}})-tr(\mathbf{\tilde{A}}) \\
			&=tr(\mathbf{\tilde{D}}^{-1}\mathbf{\tilde{A}}^2)-tr(\mathbf{\tilde{A}}) \\
			&=\sum_{i=1}^{N}(\frac{d_i+1}{d_i})-N \\
			& =\sum_{i=1}^{N}\frac{1}{d_i}>0.
	\end{align*}
	Here the second last equation is because that from the diagonal elements of $\mathbf{\tilde{A}}^2$ is the paths number from node $v_i$ to itself, apparently that in a graph with self-loops there are $d_i+1$ paths, where $d_i$ is the degree of node $v_i$. For an identity feature matrix $\mathbf{X}$, we have $\mathbf{X}^T(\mathbf{L}-\mathbf{L}_{2})\mathbf{X}$ similar to $(\mathbf{L}-\mathbf{L}_{2})$, so that $tr(\mathbf{X}^T(\mathbf{L}-\mathbf{L}_{2})\mathbf{X})=tr(\mathbf{L}-\mathbf{L}_{2})>0$.  
\end{proof}
\end{corollary}

From the above corollary, we see that the trace of $\mathbf{X}^T\mathbf{L}\mathbf{X}$ is larger than $\mathbf{X}^T\mathbf{L}_{2}\mathbf{X}$, which indicates the second-order filter smoothes the local structure. 

\subsubsection{Overall Loss Function}
Since the natural graphs always exhibit sparsity and low-rank properties \cite{jin2020graph}, we also add sparsity and low-rank regularization terms on structure learning. And we could also add the feature smoothness term when features are available. So that the overall loss function could be formulated by:
\begin{align*}
		\mathcal{L} = &\alpha\|\mathbf{S}\|_1 + \beta\|\mathbf{S}\|_* + \sum_{k=1}^{K}\eta_k\|\mathbf{\hat{S}} - \mathbf{\hat{A}}^k \|_F^2 \\
		&+ \lambda tr((\mathbf{X}^T\mathbf{\hat{L}}\mathbf{X}) + \mathcal{L}_{GNN},
\end{align*}
where the term $\mathcal{L}_{GNN}$ is the classification loss of the GNN such as cross entropy, $\|\cdot\|_*$ is the nuclear norm. We iteratively optimize the parameters of GNN and the learned adjacency matrix $\mathbf{S}$. The optimization algorithm is shown in Algorithm \ref{optimization}, where we use proximal operator for the optimization of $l_1$ norm and nuclear norm \cite{beck2009fast, richard2012estimation}, $P(\mathbf{S})$ is a projection function that project $\mathbf{S}_{ij}<0$ to $0$ and $\mathbf{S}_{ij}>1$ to $1$.

\begin{algorithm}[h]
	\caption{Optimization Algorithm.}
	\noindent\textbf{Input}: Adjacency matrix $\mathbf{A}$, feature matrix $\mathbf{X}$, labels $\mathbf{Y}_L$, hyper-parameters $\alpha$, $\beta$, $\lambda$, $\tau$, $\eta$, Learning rate $\mu$, $\mu^\prime$\\
	\noindent\textbf{Output}: Learned adjacency matrix $\mathbf{S}$, GNN parameters $\theta$.
	\begin{algorithmic}[1]
		\STATE Initialize $\mathbf{S}\leftarrow\mathbf{A}$
		\STATE Randomly initialize $\theta$
		\WHILE{stopping condition is not met}
		\STATE $\mathbf{S}\leftarrow\mathbf{S}-\mu\nabla_{\mathbf{S}}(\sum_{k=1}^{K}\eta_k\|\mathbf{\hat{S}} - \mathbf{\hat{A}}^k \|_F^2 + \lambda tr((\mathbf{X}^T\mathbf{\hat{L}}\mathbf{X}) + \mathcal{L}_{GNN})$
		\STATE $\mathbf{S}\leftarrow\mathbf{U}diag((\sigma_i-\beta)_+))_i\mathbf{V}^T$
		\STATE $\mathbf{S}\leftarrow sgn(\mathbf{S})\odot(|\mathbf{S}|-\alpha)_+$
		\STATE $\mathbf{S}\leftarrow P(\mathbf{S})$
		\FOR{$i=1$ to $\tau$}
		\STATE $\theta\leftarrow\theta-\mu^\prime\nabla_\theta\mathcal{L}_{GNN}$
		\ENDFOR
		\ENDWHILE
		\RETURN $\mathbf{S}$, $\theta$
	\end{algorithmic}
	\label{optimization} 
\end{algorithm}

\section{Experiments}
In this section, we evaluate the effectiveness of our proposed method against different types of adversarial attacks. We compare our method with several state-of-the-art methods, and analyze the parameter sensitivity of our method.

\subsection{Experimental Settings}

\subsubsection{Datasets}
We evaluate our method on three benchmark datasets, i.e., Cora, Citeseer and Polblogs, as shown in Table \ref{datasets}. Specifically, Cora and Citeseer are citation networks and they all have features, while Polblogs is a web network dataset whose features are not available.

\begin{table}[!ht]
	\centering
	\setlength{\tabcolsep}{2.3mm}{
		\begin{tabular}{c|cccc}
			\toprule[1.25pt]
			& \textbf{Nodes} & \textbf{Edges} & \textbf{Features} & \textbf{Classes}\\
			\midrule
			\textbf{Cora} & 2,708 & 5,429 & 1,433 & 7\\
			\textbf{Citeseer} & 3,327 & 4,732 & 3,703 & 6\\
			\textbf{Polblogs} & 1,222 & 16,714 & - & 2\\
			\bottomrule[1.25pt]
	\end{tabular}}
	\caption{Description of datasets used for node classification.}
	\label{datasets}
\end{table}

\subsubsection{Implementation Details}
To validate the effectiveness of our method, we compare it with several state-of-the-art defense methods. Our experiments are conducted based on the adversarial attack repository DeepRobust \cite{li2020deeprobust}.
Following the classical semi-supervised classification setting, we randomly choose $10\%$ of the nodes for training, $10\%$ for validation and $80\%$ for testing for each graph. For each experiment, we run 10 times and report the mean performance and variance of each method. We adopt the default parameter settings for the baseline methods. We use $K=3$ for all the datasets for implementation.

\begin{table*}[htbp]
	\centering
	\setlength{\tabcolsep}{2.3mm}{
		\begin{tabular}{c|c|ccccccc}
			\toprule[1.25pt]
			\textbf{Dataset} & \textbf{Ptb Rate} & \textbf{GCN} & \textbf{GAT} & \textbf{RGCN} & \textbf{Jaccard} & \textbf{ProGNN} & \textbf{ElasticGNN} & \textbf{Ours}\\
			\midrule
			\multirow{6}{*}{\textbf{Cora}} 
			& 0.00 & $83.5\pm0.4$ & $84.0\pm0.7$ & $83.1\pm0.4$ & $82.1\pm0.5$ & $83.4\pm0.5$ & $\mathbf{85.8\pm0.4}$ & $83.4\pm0.5$\\
			& 0.05 & $76.6\pm0.8$ & $80.4\pm0.7$ & $77.4\pm0.4$ & $79.1\pm0.6$ & $82.8\pm0.4$ & $82.3\pm1.1$ & $\mathbf{82.8\pm0.4}$\\
			& 0.10 & $70.4\pm1.3$ & $75.6\pm0.6$ & $72.2\pm0.4$ & $75.2\pm0.8$ & $79.0\pm0.6$ & $78.8\pm1.7$ & $\mathbf{79.3\pm1.3}$\\
			& 0.15 & $65.1\pm0.7$ & $69.8\pm1.3$ & $66.8\pm0.4$ & $71.0\pm0.6$ & $76.4\pm1.3$ & $77.2\pm1.6$ & $\mathbf{77.9\pm1.4}$\\
			& 0.20 & $59.6\pm2.7$ & $59.9\pm0.9$ & $59.3\pm0.4$ & $65.7\pm0.9$ & $73.3\pm1.6$ & $70.4\pm1.3$ & $\mathbf{73.3\pm1.4}$\\
			& 0.25 & $47.5\pm2.0$ & $54.8\pm0.7$ & $50.5\pm0.8$ & $60.8\pm1.1$ & $69.7\pm1.7$ & $-$ & $\mathbf{71.5\pm1.5}$\\
			\midrule
			\multirow{6}{*}{\textbf{Citeseer}} 
			& 0.00 & $72.0\pm0.6$ & $73.3\pm0.8$ & $71.2\pm0.8$ & $72.1\pm0.6$ & $73.3\pm0.7$ & $\mathbf{73.8\pm0.6}$ & $73.2\pm0.7$\\
			& 0.05 & $70.9\pm0.6$ & $72.9\pm0.8$ & $70.5\pm0.4$ & $70.5\pm1.0$ & $73.1\pm0.3$ & $73.3\pm0.6$ & $\mathbf{73.3\pm0.7}$\\
			& 0.10 & $67.6\pm0.9$ & $70.6\pm0.5$ & $67.7\pm0.3$ & $69.5\pm0.6$ & $72.5\pm0.8$ & $72.4\pm0.9$ & $\mathbf{72.6\pm0.3}$\\
			& 0.15 & $64.5\pm0.1$ & $69.0\pm1.1$ & $65.7\pm0.4$ & $66.0\pm0.9$ & $72.0\pm1.1$ & $71.3\pm1.5$ & $\mathbf{72.2\pm0.6}$\\
			& 0.20 & $62.0\pm3.5$ & $61.0\pm1.5$ & $62.5\pm1.2$ & $59.3\pm1.4$ & $70.0\pm2.3$ & $64.9\pm1.0$ & $\mathbf{71.5\pm0.7}$\\
			& 0.25 & $56.9\pm2.1$ & $61.9\pm1.1$ & $55.4\pm0.7$ & $59.9\pm1.5$ & $69.0\pm2.8$ & $-$ & $\mathbf{70.8\pm1.6}$\\
			\midrule
			\multirow{6}{*}{\textbf{Polblogs}} 
			& 0.00 & $95.7\pm0.4$ & $95.4\pm0.2$ & $95.2\pm0.1$ & $-$ & $93.2\pm0.6$ & $\mathbf{95.8\pm0.3}$ & $94.2\pm1.3$\\
			& 0.05 & $73.1\pm0.8$ & $83.7\pm1.5$ & $74.3\pm0.2$ & $-$ & $93.3\pm0.2$ & $83.0\pm0.3$ & $\mathbf{93.9\pm2.8}$\\
			& 0.10 & $70.7\pm1.1$ & $76.3\pm0.9$ & $71.0\pm0.3$ & $-$ & $89.4\pm1.1$ & $81.6\pm0.3$ & $\mathbf{90.0\pm3.6}$\\
			& 0.15 & $65.0\pm1.9$ & $68.8\pm1.1$ & $67.3\pm0.4$ & $-$ & $86.0\pm2.2$ & $78.7\pm0.5$ & $\mathbf{88.2\pm5.2}$\\
			& 0.20 & $51.3\pm1.2$ & $51.5\pm1.6$ & $59.9\pm0.3$ & $-$ & $79.6\pm5.7$ & $77.5\pm0.2$ & $\mathbf{86.8\pm6.2}$\\
			& 0.25 & $49.2\pm1.4$ & $51.2\pm1.5$ & $56.0\pm0.6$ & $-$ & $63.2\pm4.4$ & $-$ & $\mathbf{83.0\pm5.0}$\\
			\bottomrule[1.25pt]
	\end{tabular}}
	\caption{Experiment results of node classification tasks against Metattack.}
	\label{performance}
\end{table*}

\subsection{Defense Performance}

We conduct experiments on three attack settings, including non-targeted attack, targeted attack and random attack.

\begin{itemize}
	\item{Non-Targeted Attack.} It considers the whole graph and aims to degrade the overall performance. We adopt a recent state-of-the-art attack, Metattack \cite{zugner2018adversarial_}.
	\item{Targeted Attack.} It aims to attack  specific nodes. We adopt a recent state-of-the-art attack, Nettack \cite{zugner2018adversarial}. 
	\item{Random Attack.} It randomly perturbs the graph structure, which injects a random noise.
\end{itemize}

\begin{figure*}[!ht]
	\centering
	\includegraphics[width=\linewidth]{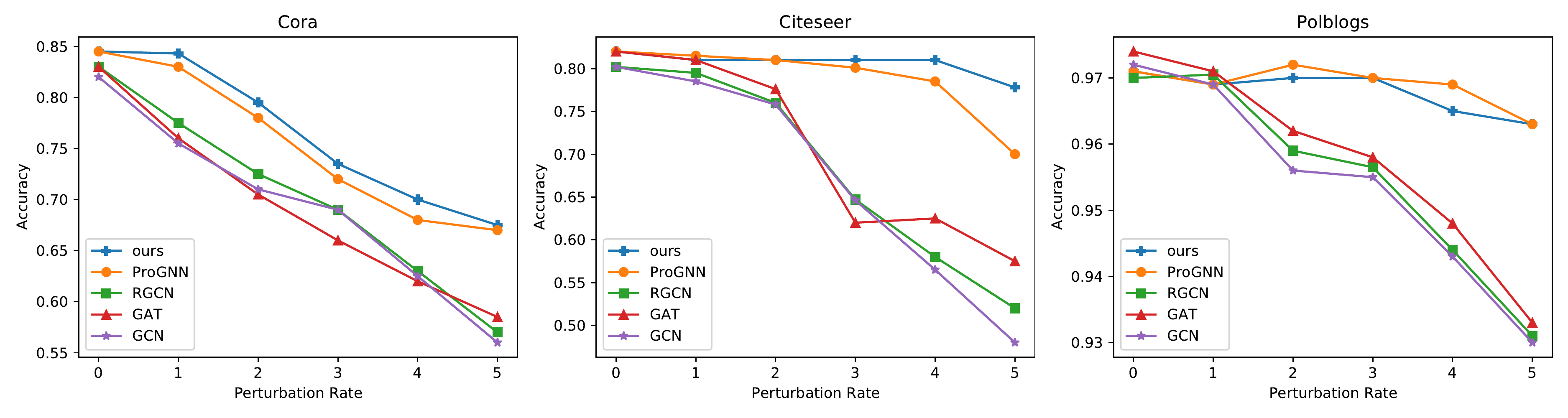}
	\caption{The classification accuracy against Nettack.}
	\label{nettack}
\end{figure*}

\begin{figure*}[!ht]
	\centering
	\includegraphics[width=\linewidth]{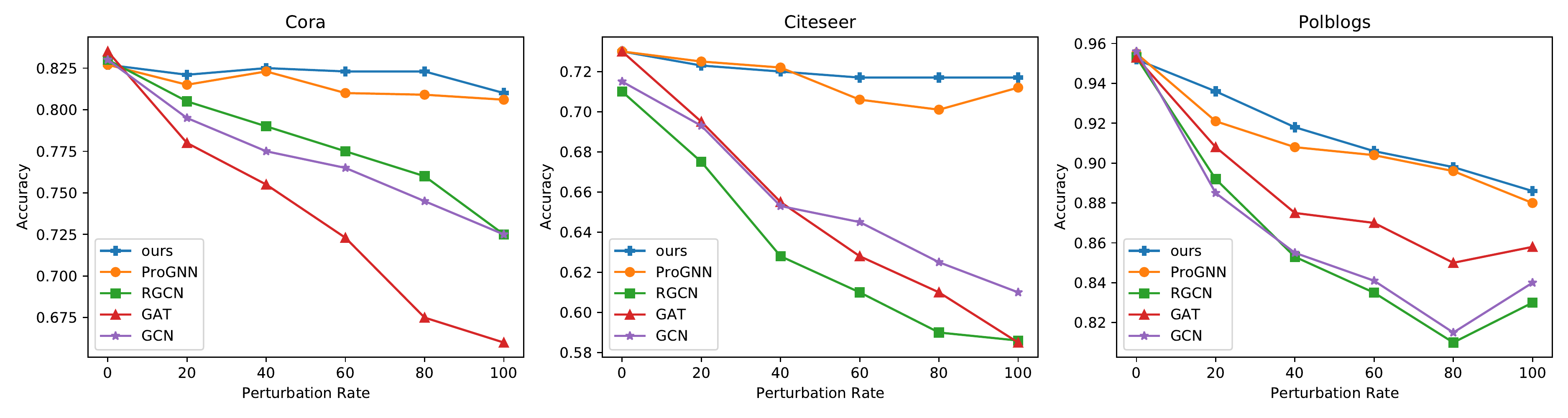}
	\caption{The classification accuracy against random attack.}
	\label{random}
\end{figure*}

\begin{figure*}[!ht]
	\centering
	\includegraphics[width=\linewidth]{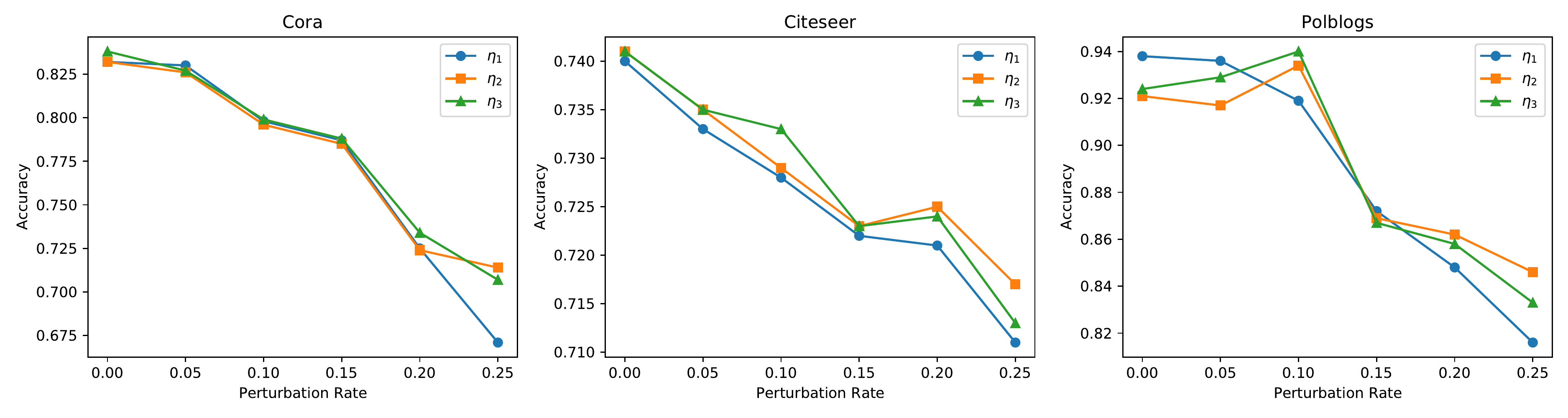}
	\caption{The classification accuracy of different hyper-parameters $\eta$ with the increase of perturbation rate.}
	\label{parameter}
\end{figure*}

\begin{figure*}[!ht]
	\centering
	\includegraphics[width=\linewidth]{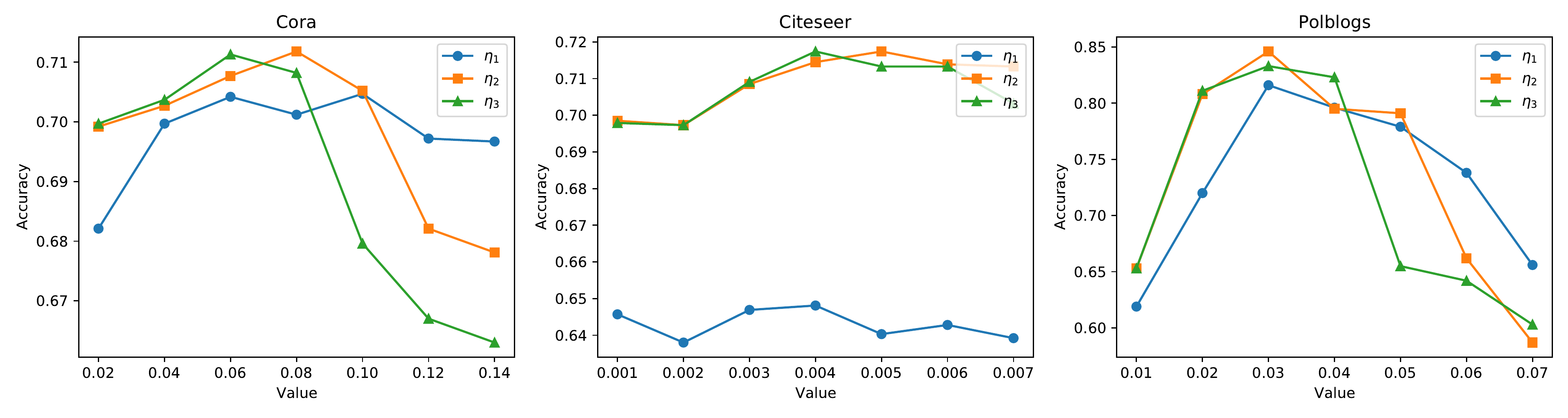}
	\caption{The parameter sensitivity of each hyper-parameter $\eta$ on three datasets.}
	\label{parameter2}
\end{figure*}

\subsubsection{Non-targeted Attack} 

We first evaluate the performance of each method against non-targeted attack. In this experiment, we adopt the Metatack for implementation, and keep all the parameter settings in original paper. To test the performance under different degrees of perturbation, we vary the perturbation rate from $0$ to $25\%$ with the step size of $5\%$. For each experiment, we run $10$ times with different seeds and report the mean accuracy and variance of each method, which is shown in Table \ref{performance}. We highlight the best performance in bold under each degree of perturbation rate.
From this table, we could make the following observations:

\begin{itemize}
	\item Our method constantly outperforms other methods under different attack degrees. For example, on each dataset, our method have different degrees of improvement compared to ProGNN. And in larger perturbation rate, our method have more distinctive improvement. Compared to vanilla GCN, our method improves $24.0\%$, $18.9\%$ and $33.8\%$ on cora, citeseer and polblogs respectively.
	\item On datasets without features, our method improves more than the baseline methods. For example, on polblogs, our method improves rather greatly compared to these methods. Especially under $25\%$ perturbation rate, our method improves $19.8\%$ compared to baselines.
\end{itemize}

\subsubsection{Targeted Attack}

We then evaluate the performance of each method against targeted attack. In this experiment, we adopt the Nettack for attack implementation and use the default parameter settings. We vary the number of perturbations on every targeted node from 1 to 5 with the step size of 1, and the nodes with degree larger than $10$ are choosed as targets. 
As shown in Figure~\ref{nettack}, we find that our method out performs the baseline methods under different degree of attacks. Moreover, compared to baselines, our method reduces the decline rate of performance with the increase of perturbation numbers. \\

\subsubsection{Random Attack}

We evaluate the performance of each method against random attack. In this experiment, we vary the random noise from $0$ to $100\%$ with the step size of $20\%$. As shown in Figure~\ref{random}, we find that our method also performs well under random attack. Still, our method performs better under larger perturbation rates, which indicates that our method can better defense against random noise.

\subsection{Parameter Analysis}
In this part, we explore the sensitivity of hyper-parameters of each order $\eta_k$ for our method. We first set the $\eta_k$ fixed and see how well they performed with the change of perturbation rate. We then set the perturbation rate fixed and check the sensitivity of each hyper-parameter. 

\subsubsection{Parameter Effect}
In this experiment, we solely use one order of regularization term, and see how they affect the performance of our method. The experimental result is shown in Figure~\ref{parameter}.
As shown in Figure~\ref{parameter}, we can observe that when the perturbation rate is small, the low-order structure hyper-parameter $\eta_1$ contributes more for the classification performance. As the perturbation rates increases, the effect of $\eta_1$ drops sharply, and the high-order structure hyper-parameter $\eta_2$ and $\eta_3$ gradually performs better than the low-order structure. For the high-order parameter, $\eta_3$ performs better when the perturbation is at a small level, while $\eta_2$ has better performance when the perturbation is relatively large.
\subsubsection{Parameter Sensitivity}
We then analyze the sensitivity of different hyper-parameters. In this part, we set the perturbation rate fixed, and analyze the performance by adjusting the value of hyper-parameters. For illustration, we use $0.25$ as the perturbation rate for all datasets, and tune the hyper-parameter to see the performance. The result is shown in Figure~\ref{parameter2}.

As shown in Figure~\ref{parameter2}, we could observe that the accuracy of can be boosted when choosing appropriate values for all the hyper-parameters. For all of these three datasets, $\eta_2$ always get the best performance; $\eta_3$ is more sensitive to the value of hyper-parameter; $\eta_1$ seems contribute less at a high-level perturbation compared to the other two parameters. Specifically, $\eta_1$ seems to help very little on Citeseer dataset at a high perturbation level, while $\eta_2$ and $\eta_3$ have similar effects.

\section{Conclusion}
Graph Neural Networks are vulnerable to the adversarial attack, where a small structure perturbation can fool GNNs into making wrong predictions. To improve the robustness of GNNs, we analyze the adversarial attack problem from the perspective of feature smoothing. We prove that high-order graph is smooth filter, which can be used to defense the adversarial attacks on graph. Therefore, we propose a novel structure learning method which explores the high-order structure of graph to help the learning process. Extensive experiments on defensing graph adversarial attacks show that our method can effectively improve the robustness of GNNs, and the proposed method outperforms recent state-of-the-art methods with a clear margin.

\bibliographystyle{abbrv}
\bibliography{main}

\end{document}